\documentclass[conference]{IEEEtran}
\IEEEoverridecommandlockouts

\usepackage{amsmath,amssymb,amsfonts}
\usepackage{booktabs}
\usepackage{cite}
\usepackage{graphicx}
\usepackage{hyperref}
\usepackage{textcomp}
\usepackage{xcolor}

\newcommand{\method}{BabySeg}
\newcommand{\website}{\href{https://w3id.org/babyseg}{w3id.org/babyseg}}
\hypersetup{
    colorlinks=true,
    citecolor=blue,
    pdftitle={\method{}},
}

\begin{document}

\title{
    \parbox{\textwidth}{
        \centering
        Deep infant brain segmentation \\
        from multi-contrast MRI
    }
    \thanks{
        The authors are grateful for funding from the National Institute of Biomedical Imaging and Bioengineering (R01 EB033773), the National Institute of Child Health and Human Development (R00 HD101553, R01 HD099846, R01 HD109436), the National Institute on Aging (R01 AG064027), the National Cancer Institute (R01 CA255479), the National Institute of Neurological Disorders and Stroke (U24 NS135561), and computational hardware provided by the Massachusetts Life Sciences Center.
    }
    \thanks{
        *~Equal contribution.
    }
}

\author{
    \IEEEauthorblockN{Malte Hoffmann\rlap{\textsuperscript{1--3}}}
    \IEEEauthorblockA{\href{mailto:mhoffmann@mgh.harvard.edu}{mhoffmann@mgh.harvard.edu}}
\and
    \IEEEauthorblockN{Lilla Z\"{o}llei\rlap{\textsuperscript{1--3,*}}}
    \IEEEauthorblockA{\href{mailto:lzollei@mgh.harvard.edu}{lzollei@mgh.harvard.edu}}
\and
    \IEEEauthorblockN{Adrian V.~Dalca\rlap{\textsuperscript{1--4,*}}}
    \IEEEauthorblockA{\href{mailto:adalca@mit.edu}{adalca@mit.edu}}
}

\makeatletter
\g@addto@macro\@maketitle{
    \centering
    \small\it
    $^1$~Athinoula A.\ Martinos Center for Biomedical Imaging \quad
    $^2$~Department of Radiology, Massachusetts General Hospital \\
    $^3$~Department of Radiology, Harvard Medical School \quad
    $^4$~Computer Science \& Artificial Intelligence Laboratory, MIT
    \vspace{\baselineskip}
}
\makeatother

\maketitle

\begin{abstract}
Segmentation of magnetic resonance images (MRI) facilitates analysis of human brain development by delineating anatomical structures. However, in infants and young children, accurate segmentation is challenging due to development and imaging constraints. Pediatric brain MRI is notoriously difficult to acquire, with inconsistent availability of imaging modalities, substantial non-head anatomy in the field of view, and frequent motion artifacts. This has led to specialized segmentation models that are often limited to specific image types or narrow age groups, or that are fragile for more variable images such as those acquired clinically. We address this method fragmentation with \method{}, a deep learning brain segmentation framework for infants and young children that supports diverse MRI protocols, including repeat scans and image types unavailable during training. Our approach builds on recent domain randomization techniques, which synthesize training images far beyond realistic bounds to promote dataset shift invariance. We also describe a mechanism that enables models to flexibly pool and interact features from any number of input scans. We demonstrate state-of-the-art performance that matches or exceeds the accuracy of several existing methods for various age cohorts and input configurations using a single model, in a fraction of the runtime required by many existing tools.
\end{abstract}


\begin{figure}[b]
    \centering
    \includegraphics[width=\linewidth]{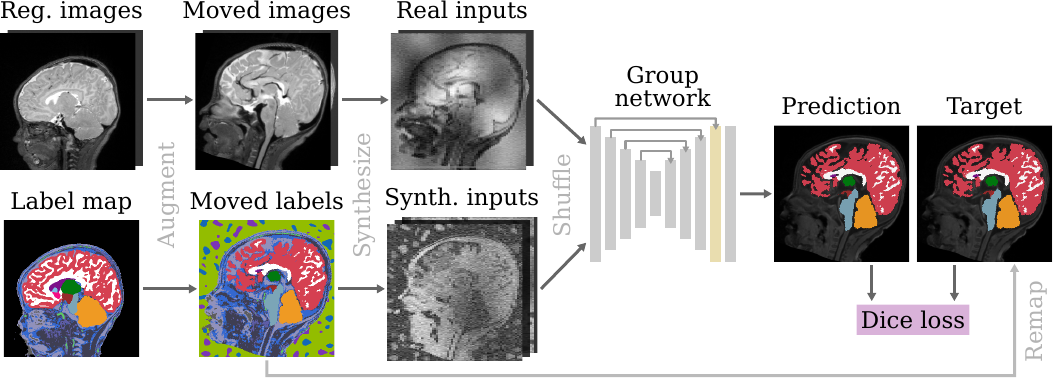}
    \caption{\method{} framework. At every step, we sample images and a label map from a subject. First, we randomly move and deform these inputs, adding synthetic background structures to the label map. Second, we synthesize training images and apply a series of corruptions. Third, the \method{} U-Net predicts a single segmentation map from these inputs.}
    \label{fig:strategy}
\end{figure}

\section{Introduction}

Anatomical segmentation of pediatric neuroimaging data is essential for automated morphometry, enabling applications from characterizing individual brain growth and maturation, understanding developmental disorders, supporting population-level interpretation, to informing dependent analyses such as those fitting cortical surface meshes\cite{dai2013ibeat,wang2023ibeat}.

Magnetic resonance imaging (MRI) is the preferred modality for many of these applications due to the ability to resolve the tissues of the brain with exquisite contrast. However, automated segmentation is more difficult in neuroimaging of infants and young children than in adults.

Although there are infant-specific coils and MRI scanners, research studies and clinical examinations still predominantly rely on hardware developed for general populations\cite{ghotra2021size}. Because of the smaller size of children, the field of view (FOV) of pediatric scans typically encompasses substantial anatomy beyond the head, such as part of the torso, the upper, and even lower extremities. Depending on their proximity to the receiver coils, these structures can appear more prominent than the brain anatomy of interest.

While the major cortical folding patterns are already in place at full-term birth, around 40 weeks gestation, the brain undergoes substantial development in early childhood, with its volume increasing two-fold during the first two years\cite{dubois2021mri}. 
 
\begin{figure}
    \centering
    \includegraphics[width=\linewidth]{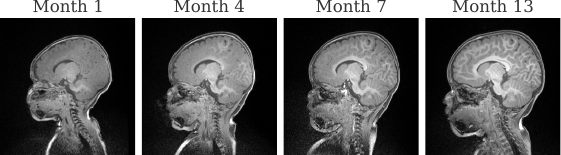}
    \caption{T1-weighted contrast. At birth, the cerebral cortex appears bright and the white matter dark. By year 1, this pattern reverses. During the inversion phase, overall contrast is low and may vary within a single scan. At month 4, for example, the shown subject has bright anterior and dark posterior cortex.}
    \label{fig:contrast}
\end{figure}

Progressive myelination profoundly impacts the imaging contrast, as captured in Fig.~\ref{fig:contrast}. In T1-weighted (T1w) neonatal MRI, the cerebral cortex appears bright and the white matter (WM) appears dark. After the first year of postnatal development, however, the cortex is dark and the WM is bright. Consequently, the contrast flips during a period characterized by low contrast overall, termed ``isointense phase'', which generally occurs between 6 and 8 months postnatally\cite{zhang2015deep}. As myelination progresses non-uniformly, a single scan may show mixed contrast, for example, with cortex appearing bright in the anterior and dark in the posterior regions of the brain.

T2-weighted (T2w) MRI exhibits a similar but opposite soft-tissue contrast compared to T1w sequences, and many pediatric studies acquire both modalities\cite{howell2019unc,edwards2022developing}.
Unfortunately, subject motion during the imaging session, which can require tens of minutes, introduces artifacts that compromise image interpretation\cite{zaitsev2015motion}. Pediatric subjects often struggle to remain still. In some cases, persistent motion prompts technologists to prematurely end a session, preventing collection of the full set of sequences required for the study\cite{afacan2016evaluation,jaimes2016strategies}.

Overall, pediatric neuroimaging is inherently complex. Achieving reliable and consistent segmentation is challenging and requires dedicated algorithms capable of accommodating the full breadth of developmental and imaging variability, to avoid a proliferation of narrow tools that each only cater to a specific scenario. While pediatric segmentation tools exist, many are limited to narrow age cohorts and modalities, or segment coarse tissue classes. Others impose substantial computational demands and remain inefficient even with those demands met, hindering large-scale deployment\cite{dai2013ibeat,wang2023ibeat}.

\subsection{Contributions}

These limitations highlight the need for a unified, scalable, and efficient approach. We introduce \method{}, a contrast-independent segmentation framework for infants and young children, making several contributions:

\begin{itemize}
    \item We present a training data engine that builds on domain randomization to combine synthetic \textit{and} real data, enabling generalization to unseen image types while boosting performance for features that are hard to synthesize.

    \item We demonstrate consistent state-of-the-art segmentation accuracy, from extremely preterm birth across the first five postnatal years of development.

    \item We introduce a convolutional mechanism that enables segmentation from a flexible number of input modalities, by pooling and interaction of intermediate features.

    \item We achieve runtimes that undercut baseline methods by 11- to 24-fold for T1w-T2w segmentation on limited hardware, with additional speed-ups if a GPU is available.
\end{itemize}

Overall, \method{} tackles longstanding challenges in infant brain segmentation. Model code and weights are freely available at \website{}. The following section will situate our framework within the broader methodological landscape.

\section{Related work}

Decades of method development have led to adult-brain segmentation techniques\cite{van2003unifying,ashburner2005unified,fischl2012freesurfer,billot2023synthseg}. Because these generally only apply to children five years and older\cite{ghosh2010evaluating}, several approaches specifically target infant populations instead.

\subsection{Infant brain segmentation}

Traditional, optimization-based segmentation methods rely on atlases: high-quality reference images with label annotations representing a population\cite{evans19933d,mazziotta1995probabilistic,sunkin2012allen}. These techniques use registration to estimate the spatial mapping between the atlas and the image. Given the transform, methods transfer the prior knowledge from the atlas to the presented anatomy\cite{bajcsy1983computerized,collins1995automatic,iosifescu1997automated,dawant2002automatic}.
In early childhood, combining information from multiple atlases is often helpful for covering a spectrum of development. While it is possible to select an appropriate atlas based on age passed as a user input\cite{sanchez2012neurodevelopmental,dai2013ibeat}, many approaches automatically select an atlas or fuse labels across multiple atlases\cite{sabuncu2010generative,wang2015links,zollei2020infant}.

Deep-learning (DL) approaches train deep neural networks to predict segmentation maps directly from input images, using a training set of examples\cite{zhang2015deep,wang2023ibeat,henschel2024vinna,hendrickson2025bibsnet}. Fully convolutional architectures such as the U-Net\cite{ronneberger2015u} are widely adopted, as their inductive biases (local connectivity, translation equivariance, and weight sharing) yield parameter- and data-efficient segmentation models. More recent work demonstrates that other architectures, like vision transformers, can achieve similar state-of-the-art results\cite{zhang2022tw}. 

DL methods for infant MRI rely on training sets that span the target cohort. However, public datasets are limited and their labels often incompatible, leading many approaches to concentrate on narrow developmental windows, such as the neonatal\cite{shen2023automatic,henschel2024vinna} or isointense phases\cite{kumar2018infinet,wang2019benchmark}. Brain growth reinforces this constraint: because most methods conform images to a common voxel size, selecting that size becomes a trade-off. Small voxels are necessary to resolve fine neonatal anatomy, yet they inflate memory and computational demands when the FOV must accommodate larger brains in later postnatal years. Consequently, tools tend to become even more narrowly scoped and fragmented.

In contrast, \method{} trains with small voxels and a medium FOV; its extensive spatial augmentation enables accurate inference with a FOV larger than during training, accommodating the brain sizes of children up to several years old.

\subsection{Domain randomization}

Standard learning-based models only work well on the type of data available at training.
Domain randomization, an emerging strategy to build robustness to input modalities, involves training networks with synthetic images that vary widely beyond features realizable with existing imaging technology\cite{billot2020learning,gopinath2024synthetic,hoffmann2025domain}. The core idea is to expose models to vast variability of image types, encouraging generalization beyond a specific set of annotated modalities usually available during training.
A procedural generative model samples from a set of precomputed label maps and synthesizes training images from them, generally by assigning a random intensity to each anatomical structure. It then corrupts the noise-free images with artifacts, broadly inspired by nuisance signals and undesirable effects in the image formation of MRI, such as intensity noise, contrast saturation, geometric distortions, and electromagnetic field inhomogeneities. Domain randomization typically employs simple models for these effects, deliberately randomizing their strength well above the levels seen in practice. This strategy has led to robust models for tasks like segmentation, registration, image-to-image translation, and structure outline reconstruction\cite{billot2023synthseg,dey2025learning,hoopes2022synthstrip,kelley2024boosting,fu2025}.

The scarcity of publicly available pediatric data for training has motivated several recent pediatric techniques to adopt randomization\cite{hendrickson2025bibsnet,shang2022learning,zalevskyi2024maximizing,shang2025towards}. Unfortunately, they tend to produce inaccurate segmentations for some of the datasets we test, and few of them make usability and runtime a priority.

\subsection{Variable input availability}

A common approach to handling multiple modalities in medical image analysis stacks the available images along the channel dimension\cite{zhang2015deep,wang2019benchmark}. This strategy ties the number of inputs to the network parameters, limits the training data to only those subjects that have the modalities envisioned, and similarly limits the utility of the network once trained.

Instead, current pipelines for multi-modal pediatric segmentation train a separate network for each input modality\cite{wang2019benchmark,hendrickson2025bibsnet,shang2022learning,shang2025towards}. Given several modalities, some methods fuse predictions across networks\cite{shang2022learning,shang2025towards}. Others train additional networks, for example, to process T1w-T2w input pairs\cite{hendrickson2025bibsnet}.

An alternative strategy for handling missing modalities is to synthesize them from the available inputs to perform segmentation with a network requiring a fixed modality set. Although approaches based on autoencoder reconstruction\cite{bachmann2022multimae} and generative adversarial learning\cite{sharma2019missing,zhang2024unified} have been explored extensively, they cannot recover information absent from the original inputs and are prone to hallucination\cite{yi2019generative}.

Recent methods can ingest a flexible number of input images by using a non-parametric mechanism, where the number of network parameters is not tied to the number of inputs. Cross-convolutional approaches leverage weight sharing to separately convolve each input in parallel, before fusing the resulting features to enable interaction across inputs\cite{czolbe2023neuralizer,butoi2023universeg,abulnaga2025multimorph}.
They have supported applications including segmentation of new structures without retraining\cite{butoi2023universeg,rakic2025pancakes}, adaptation to new image processing tasks, and one-shot atlas creation from a flexible number of images\cite{abulnaga2025multimorph}.

In this work, we describe a flexible mechanism inspired by cross convolutions, and build on domain randomization to enable integration of unseen modalities, including repeat scans within the same session, for segmentation without retraining.

\section{Proposed method}

\newcommand{\voxel}{\mathbf{v}}
\newcommand{\labels}{\mathbf{y}}
\newcommand{\image}{\mathbf{x}}
\newcommand{\network}{g}
\newcommand{\feature}{\mathbf{f}}

\begin{figure}
    \centering
    \includegraphics[width=0.55\linewidth]{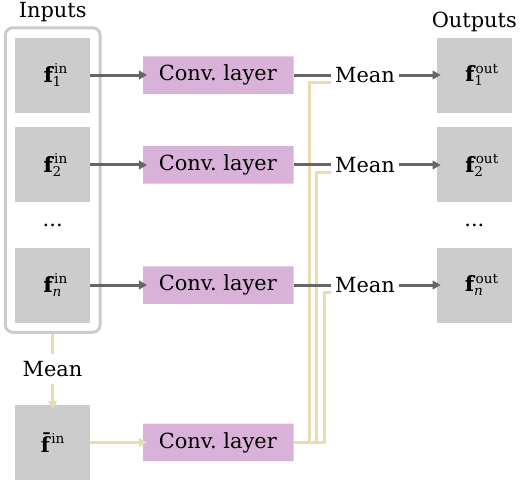}
    \caption{Group convolution. The convolutional layer supports a \textit{flexible} number of $n$ inputs by first separately convolving each group entry $\feature^\mathrm{in}_i$, where $i \in \{1, 2, ..., n\}$, as well as convolving the mean group representation $\bar{\feature}^\mathrm{in}$. It encourages interaction across group entries by averaging each with the convolved group mean to output $\feature^\mathrm{out}_i$.}
    \label{fig:layer}
\end{figure}

\begin{figure}
    \centering
    \includegraphics[width=\linewidth]{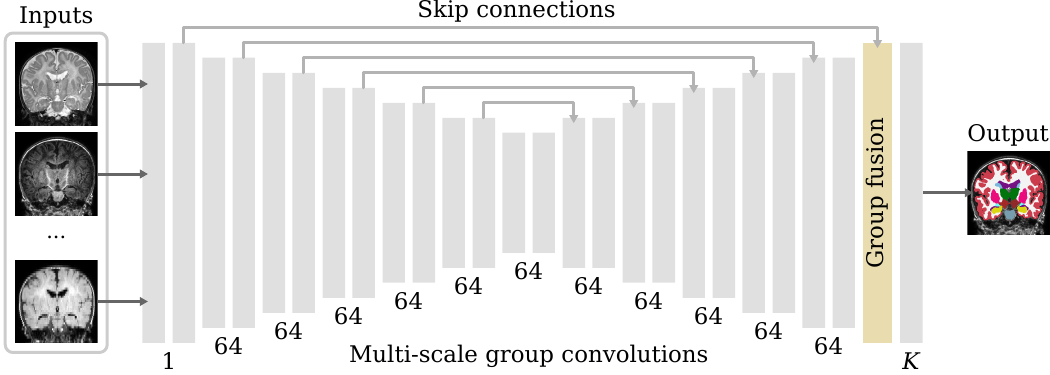}
    \caption{Group segmentation network. \method{} implements a U-Net, which pools features at multiple resolutions. Each gray layer applies a group convolution, interacting features from a \textit{flexible} number of inputs. A group fusion layer averages across groups to output a single label map delineating $K$ anatomical structures including background. Numbers indicate filter count.}
    \label{fig:network}
\end{figure}

Let $\image_i$, $i \in \{1,\dots,n\}$ be a group of $n$ spatially aligned, potentially multi-modal $d$-dimensional images that represent the same anatomy. Let $\labels$ be an associated map of probabilities $\labels(k, \voxel)$ that grid point $\voxel$ belongs to class $k \in \{1,\dots,K\}$, with $\sum_{k=1}^K{\labels(k, \voxel)} = 1$. We model a function $\network$ that predicts
\begin{equation}
    \network(\image_1,\dots,\image_n) = \labels
    \label{eq:function}
\end{equation}
in a forward pass, such that $\network$ can be applied to any number $n$ of input scans of any types and sizes.

\subsection{Method overview}

We propose a group segmentation framework, \method{}, implementing \eqref{eq:function} as a convolutional neural network $\network_\theta$ of learnable parameters $\theta$.

Fig.~\ref{fig:strategy} provides an overview of the learning strategy. At every step, we sample a pre-generated label map $\labels$ and all associated, acquired, and spatially registered images $\image_i$, $i \in \{1, \dots, n\}$, available from one imaging session in training set $\mathcal{D}$.
Group network $\network_\theta$ receives the images as an input and predicts a single one-hot probability map $\hat{\labels} = \network_\theta(\image_1,\dots,\image_n)$. The learning setup is supervised: we fit $\theta$ subject to
\begin{equation}
    \theta^* = \mathop{\arg \min}_\theta
    \;
    \mathbb{E}_\mathcal{D}
    \Big[
    \mathcal{L}(\network_\theta(\image_1, \dots, \image_n), \labels)
    \Big],
\end{equation}
minimizing a criterion $\mathcal{L}$ that measures the overlap of $\hat{\labels}$ with the ground-truth labels $\labels$\cite{milletari2016v},
\begin{equation}
    \mathcal{L}(\labels, \hat{\labels}) = 1 - 
    \sum_{k=1}^{K}{
        \frac{
            2
            \sum_\voxel{\labels(k,\voxel)
            \,
            \hat{\labels}(k,\voxel)}
        }
        {
        \sum_\voxel{\big[\labels(k,\voxel)^2} +
            \hat{\labels}(k,\voxel)^2 \big]
        }
    }.
\end{equation}

This framework combines two key parts: (1) a group convolutional network architecture and (2) a data engine that drives \method{} training by generating new, artificial contrasts from label maps and acquired MRI scans.

\subsection{Segmentation model}

We introduce an architecture able to predict a label map from a group of input modalities whose size $n$ can vary during training and inference. Each network layer pools and enables interaction between features from all group entries.

\subsubsection{Group convolution}

Instead of passing group entries as channels to standard convolutions, which fixes the input channel count, we propose a flexible and efficient group convolution mechanism, illustrated in Fig.~\ref{fig:layer}. Let $*$ denote a $d$-dimensional convolution and $\feature^\mathrm{in}_i$ the input feature maps for group entry $i \in \{1,\dots,n\}$, corresponding to scan~$\image_i$. The group convolutional layer enables cross-talk between the group entries through the mean group representation $\bar{\feature}^\mathrm{in}$, and it outputs feature maps
\begin{equation}
    \feature_i^\mathrm{out} =
    0.5 \, \big(
    \mathbf{V} * \feature_i^\mathrm{in} +
    \mathbf{W} * \bar{\feature}^\mathrm{in}
    \big)
    +\mathbf{b}
\end{equation}
with
\begin{equation}
    \bar{\feature}^\mathrm{in} = \frac{1}{n}\sum_{i=1}^n{\feature^\mathrm{in}_{i}},
\end{equation}
where $\mathbf{b}$ is a learnable bias vector and $\mathbf{V}, \mathbf{W}$ are learnable kernel weights.

\subsubsection{Network architecture}

We implement a segmentation network using the well established U-Net\cite{ronneberger2015u} architecture, pooling features across multiple spatial scales with group-convolutional in place of conventional convolutional layers. As shown in Fig.~\ref{fig:network}, the network includes the usual skip concatenations from features extracted in the downsampling arm to the corresponding level in the upsampling arm.

Each group convolution returns one feature map for each input group scan $\image_i$. To output a single $K$-channel label map, we fuse features across groups by computing the mean over the nonlinearly activated outputs of the second-last layer before applying a final upsampling operation and $K$-channel group convolution. Sec.~\ref{sec:details} details the implementation.

\subsection{Data engine}

\begin{figure}
    \centering
    \includegraphics[width=\linewidth]{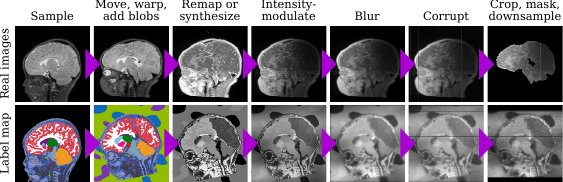}
    \caption{Data engine. First, we sample, move, and warp images and the associated label map, adding synthetic background blobs to the latter. Second, we remap intensities and generate images from the label map. Third, we apply a series of corruptions, to obtain images for training. All steps are randomized.}
    \label{fig:engine}
\end{figure}

We train jointly on synthetic and real data. This enables us to leverage the inference-time robustness provided by synthetic data, while learning precise signals from real-world scans. To this end, we introduce a procedural data engine that creates new, artificial imaging contrasts from label maps \textit{and} acquired images for training (Fig.~\ref{fig:engine}). In this section, we briefly summarize concepts from our prior work\cite{hoffmann2021synthmorph,hoffmann2024anatomy,hoopes2022synthstrip,kelley2024boosting,specktor2025singlestrip}, recently detailed in a tutorial\cite{hoffmann2025domain}, to focus on what is novel.

\subsubsection{Spatial augmentation}

Starting with a set of spatially registered images and an associated label map from the same session, we simulate a new, deformed anatomy and orientation in the FOV by applying a randomized nonlinear deformation including global translation, rotation, scaling, and shear\cite{hoffmann2021synthmorph}.
For robustness to non-head anatomy in the FOV, we replace the label map background with $n_\mathrm{blobs} \sim \mathcal{U}(0, 3)$, $n_\mathrm{blobs} \in \mathbb{N}$ geometric shape labels.
First, we draw a $d$-dimensional gradient-noise image $\image_\mathrm{blobs}$\cite{perlin1985image}.
Second, we uniformly sample a threshold $t \sim \mathcal{U}(0, 0.5)$ and set intensities $\image_\mathrm{blobs}(\voxel)$ for all grid points $\voxel$ with absolute values below $t$ to zero.
Third, we convert the thresholded image to discrete labels by sorting its grid-point intensities into $n_\mathrm{blobs}$ equally sized bins.

\subsubsection{Artificial contrasts}

From the spatially augmented inputs, we synthesize $n \sim \mathcal{U}(1, 4)$, $n \in \mathbb{N}$ images with new, artificial contrasts. The goal is to promote network robustness via exposure to broad variability in training, rather than simulating realistic neuroimages.

Assuming $\tilde{n}_\mathrm{real}$ acquired images are available for the session, we retain a subset $n_\mathrm{real} \sim \mathcal{U}(0, \min(\tilde{n}_\mathrm{real}, n))$, and separately remap the intensity distribution of each as follows. First, we min-max normalize image intensities to an 8-bit range. Second, we generate a smooth lookup table $\mathbf{L}$ from gradient noise, which associates each 8-bit value with a new, random output intensity. Third, we remap the image, interpreting its intensity values as indices into $\mathbf{L}$.

A similar operation synthesizes the remaining $n_\mathrm{synth} = n - n_\mathrm{real}$ gray-scale images from the label map $\labels$: for each of the $n_\mathrm{synth}$ outputs, we uniformly sample a random intensity value for each label $k$ defined in $\labels$ and assign that value to all grid points associated with $k$. In contrast to the remapped real scans, these images are noise free (for now), but structure intensities vary laterally. For example, the left hippocampus might be bright and the right hippocampus dark in one realization, whereas the next might reverse this contrast.

\subsubsection{Image corruptions}

Next, the data engine separately applies a series of randomized corruption steps to each of the images to yield the final data-engine outputs, used as network inputs, $\image_1, \dots, \image_n$\cite{hoffmann2025domain}.
These steps include bias modulation, partial voluming, additive Gaussian noise, gamma transformation, resolution reduction, and cropping.
We replace $n_\mathrm{slice} \sim \mathcal{U}(1, 3)$, $n_\mathrm{slice} \in \mathbb{N}$ slices along an axis with a constant, random intensity value. Additionally, we reduce the resolution along an axis to simulate thick slices often found in clinical-quality scans. We also simulate randomized skull-stripping with over- and undersegmentation, by constructing a $d$-dimensional binary brain mask $\mathbf{m}_i$ from all brain labels in the label map. We apply binary hole-filling and closing to this mask, using $n_\mathrm{dilate} \sim \mathcal{U}(0, 10)$ dilation and $n_\mathrm{erode} = n_\mathrm{dilate} - \Delta{n}_\mathrm{erode}$ erosion steps, with
\begin{equation}
    \Delta{n}_\mathrm{erode} \sim \mathcal{U}
    (\max(-n_\mathrm{dilate}, -4), \min(n_\mathrm{dilate}, 8)),
\end{equation}
where $n_\mathrm{dilate} \in \mathbb{N}$, $\Delta{n}_\mathrm{erode} \in \mathbb{Z}$, and we apply it to the image via element-wise multiplication. As each image is assigned a separate $\mathbf{m}_i$, we account for undersegmentation using an aggregate mask $\mathbf{m}$ for the label map $\labels$, where, for all grid points~$\voxel$,
\begin{equation}
    \mathbf{m}(\voxel) = \max_{i \in \{1, \dots, n\}} \mathbf{m}_i(\voxel).
\end{equation}

\begin{table}
    \centering
    \caption{Uniform domain-randomization ranges $[a, b]$ and corruption probabilities $p$. We abbreviate standard deviation (SD), full width at half maximum (FWHM), and field of view (FOV).}
    \label{tab:ranges}
    \small
    \begin{tabular}{llrrr}
    \toprule
    Parameter & Unit & $a$ & $b$ & $p$ \\
    \midrule
    Translation               & mm       & $-45$ & 45  & 1.0 \\
    Rotation                  & $^\circ$ & $-30$ & 30  & 1.0 \\
    Scaling                   & \%       & 90    & 110 & 1.0 \\
    Shear                     & \%       & 90    & 110 & 1.0 \\
    Warp displacement         & mm       & 0     & 18  & 1.0 \\
    Warp control points       & --       & 2     & 16  &     \\
    Left-right flipping       & --       & --    & --  & 0.5 \\
    Blob label count          & --       & 1     & 3   & 0.5 \\
    Blob control points       & --       & 2     & 4   &     \\
    Label intensity mean      & a.u.     & 0     & 1   &     \\
    Image channel count       & --       & 1     & 4   &     \\
    Real channel count        & --       & 1     & 2   & 0.5 \\
    Lookup control points     & --       & 2     & 8   & 0.5 \\
    Bias field drop           & \%       & 0     & 50  & 1.0 \\
    Bias field control points & mm       & 2     & 4   &     \\
    Image blurring FWHM       & mm       & 0     & 3   & 1.0 \\
    Noise intensity SD        & \%       & 0     & 10  & 1.0 \\
    Slice fill count          & --       & 1     & 3   & 0.5 \\
    Slice fill intensity      & a.u.     & 0     & 1   &     \\
    Downsampling factor       & --       & 1     & 4   & 0.5 \\
    Gamma exponent            & --       & 0.5   & 1.5 & 1.0 \\
    FOV cropping              & \%       & 0     & 33  & 0.5 \\
    Skull-stripping           & --       & --    & --  & 0.5 \\
    Skull-strip dilation      & --       & 0     & 10  & 1.0 \\
    Skull-strip erosion delta & --       & $-4$  &  8  & 1.0 \\
    Skull-strip hole filling  & --       & --    & --  & 0.5 \\
    \bottomrule
    \end{tabular}
\end{table}

\subsubsection{Randomization ranges}

The synthesis and augmentation steps have been laid out in detail\cite{gopinath2024synthetic,hoffmann2025domain}. We follow prior studies, which optimize randomization ranges for segmentation\cite{hoopes2022synthstrip,billot2023synthseg,kelley2024boosting}, and report the parameters in Tab.~\ref{tab:ranges}.

\section{Experiment}

In this section, we validate the proposed method and compare its performance to state-of-the-art models dedicated to infant brain segmentation across age cohorts.

\subsection{Data}
\label{sec:data}

We use four datasets: Baby Open Brains from 2024-10-10 (BOBs)\cite{feczko2025baby}, the Melbourne Children's Regional Infant Brain dataset (M-CRIB)\cite{alexander2017new}, release 4 of the Developing Human Connectome Project (dHCP)\cite{edwards2022developing}, and the Infant FreeSurfer dataset (IFS)\cite{zollei2020infant}.

BOBs and M-CRIB include 71 and 10 subjects, respectively, with T1w and T2w images for each. BOBs subjects are aged 1--9 months, whereas M-CRIB covers 40--43 weeks postmenstrual age at scan. We have dHCP data from 782 term and preterm newborns, some of whom participated in several imaging sessions, for a total of 709 T1w and 885 T2w images covering a period of 27--45 weeks postmenstrual age at scan.
The subset of IFS data used here comprises T1w scans from 15 subjects aged 0--5 years.

We retain 4 BOBs and 4 dHCP subjects for validation, holding out another 20 dHCP subjects for testing, spread evenly across the postmenstrual age range. The training set includes all remaining BOBs and dHCP data, whereas we use M-CRIB and IFS exclusively for out-of-distribution testing.

\begin{figure*}
    \centering
    \includegraphics[width=\linewidth]{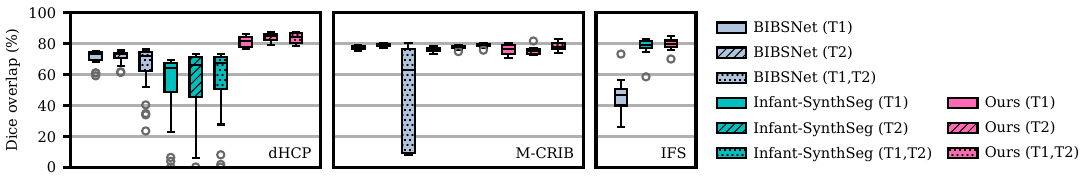}
    \caption{\method{} segmentation accuracy compared to baseline methods across test sets and input-image types, in parentheses. Higher Dice scores are better.}
    \label{fig:accuracy}
\end{figure*}

\subsection{Labels}
\label{sec:labels}

While each dataset comes with a label map for each session, the anatomical labels are not identical across datasets. To obtain a single, consistent set of anatomical structures for \method{} training, we remap the label maps to the largest possible set of 21 overlapping FreeSurfer structures\cite{fischl2012freesurfer}: cerebral white matter (WM), cerebral cortex, lateral ventricles, cerebellum, thalamus, caudate, brain stem, hippocampus, amygdala, ventral DC, and basal ganglia. Of these, only the brain stem is a unilateral structure. Combined with the background, these labels represent the $K$ segmentation labels output by \method{}.

Most of the labels included in the datasets directly translate to the \method{} labels, with two exceptions: vermis and corpus callosum (CC).
Some BOBs subjects have a cerebellar vermis label, while others have the corresponding voxels assigned to the left or right cerebellar cortex. Similarly, dHCP's CC label does not exist in BOBs; the corresponding voxels are either left or right cerebral WM. We split vermis and CC via registration to a reference annotated with BabySeg labels, followed by manual label refinement as needed.

In addition to the blobs generated by the data engine, we complete the \textit{training} label maps with a further six non-brain labels by fitting a Gaussian mixture model to non-zero T2w voxels outside the brain\cite{hoopes2022synthstrip,kelley2024boosting}. These labels are not consistent across images or subjects, but they enable synthesis of whole-head images for \method{} training.

\begin{figure}[b]
    \centering
    \includegraphics[width=\linewidth]{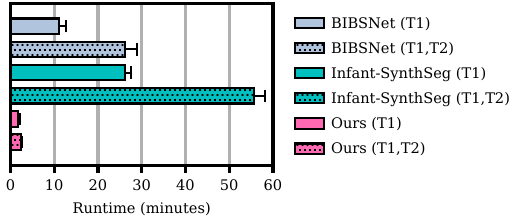}
    \caption{Mean single-threaded CPU-only runtime over $n = 10$ M-CRIB subjects by input type, in parentheses. Error bars indicate standard deviations.}
    \label{fig:runtime}
\end{figure}

\subsection{Implementation details}
\label{sec:details}

For training and internal validation, we resample images and label maps to volumes of $192 \times 192 \times 192$ isotropic 0.7-mm voxels, centering the brain with left-inferior-anterior (LIA) orientation relative to the buffer axes. We normalize each input image by clipping intensities outside the 1--99 percentile range and min-max rescaling values to $[0, 1]$.

The model implements a downsampling arm with 6 levels, each applying two convolutions with $3 \times 3 \times 3$ kernels and exponential linear unit (ELU) activation\cite{clevert2015fast}, followed by max pooling (Fig.~\ref{fig:network}). The upsampling arm mirrors this structure, with linear upsampling, skip-concatenating the downsampling features. All convolutions use 64 filters, except for the first level, which derives a single feature, and the final convolution, which outputs $K=22$ softmax-activated maps.

Training uses the Adam optimizer with a batch size of 1 and an initial learning rate of $10^{-4}$ until improvements to the validation Dice curve visually slow. We then reduce the learning rate to $10^{-5}$ and train until the curve remains visually stable, after a total of at least 1 million steps. For the baseline comparison, we select the model parameters leading to the highest accuracy over the validation images (Sec.~\ref{sec:data}). For training and inference, we use an Nvidia A100 GPU with 40 GB of memory and compare model runtimes on an Intel Xeon Silver 4114 processor with a clock rate of 2.2 GHz.

\subsection{Baseline methods}

We evaluate recent DL methods for infant-specific brain segmentation that use domain-randomization techniques similar to \method's. First, we test Infant-SynthSeg\cite{shang2022learning,shang2025towards}. This method trains separate single-input U-Nets for T1w and T2w images, fusing their predictions after removing connected components of 15 or fewer voxels by assigning the label with the largest posterior across its neighbors. Infant-SynthSeg aims to build robustness to cross-structure contrast variability in the first postnatal months by synthesizing T1w and T2w images for training from label maps with manually subdivided anatomical structures and assigning variable intensities.

We also evaluate BIBSNet\cite{hendrickson2025bibsnet}, a segmentation pipeline running nnU-Net\cite{isensee2021nnu} models trained on a mixture of real and synthetic T1w, T2w, or both T1w and T2w images. By selecting a model for inference depending on the inputs provided, BIBSNet avoids the label fusion step. While Infant-SynthSeg assumes registered input T1w-T2w pairs, BIBSNet includes an internal registration step. Running these methods with varied input contrasts provides a total of six baseline setups for comparison.

\subsection{Experimental setup}

For each method, we separately test segmentation using each possible input configuration for an imaging session: T1w only, T2w only, and both.
Since the baseline methods segment slightly different structures than \method{}, the method comparison measures accuracy as mean Dice scores over the bilateral labels of Sec.~\ref{sec:labels} adjusted as follows. First, we merge the cerebellar cortex, WM, and vermis, if segmented, into a single cerebellar label. Second, we fuse the pallidum, putamen, and nucleus accumbens to form a basal ganglia label. Third, we combine the choroid plexus and inferior lateral ventricles into a single, lateralized ventricular label. 
For a fair comparison, we correct alignment errors between T1w and T2w scans using registration\cite{hoffmann2024anatomy,fu2025}.

\subsection{Results and discussion}

\begin{figure}
    \centering
    \includegraphics[width=\linewidth]{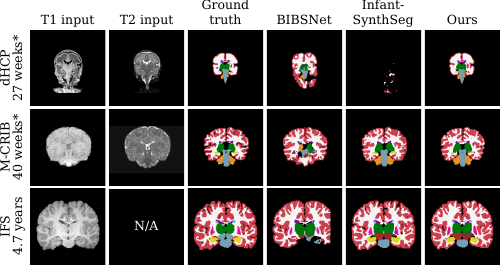}
    \caption{Segmentation examples across methods. Each row shows a subject from a different test set. Asterisks represent postmenstrual age.}
    \label{fig:examples}
\end{figure}

Fig.~\ref{fig:examples} visualizes sample segmentations for each method, evaluated using all available input contrasts. Qualitatively, \method{} generalizes to severely preterm neonates scanned at 26 weeks postmenstrual age, while the baseline methods substantially over- or undersegment. \method{} and Infant-SynthSeg accurately segment the oldest IFS subject scanned at 4.7 years. In contrast, BIBSNet's prediction undersegments, likely because the size of the brain exceeds the FOV covered by the template it uses for internal registration.

In Fig.~\ref{fig:accuracy}, we quantitatively compare segmentation accuracy for each input configuration. \method{} generally ranks among the two best-performing methods and, crucially, avoids low-accuracy cases of mean Dice scores substantially less than 70 points, which each of the baselines yield in several test settings. As expected, \method{} reaches its highest Dice scores for the held-out in-distribution dHCP test set. However, its performance remains high out of distribution, such as for the IFS subjects, whose age at scan exceeds the oldest training subjects by up to 4 years.
Providing both T1w and T2w inputs slightly improves median Dice scores across dHCP and M-CRIB except for BIBSNet. BIBSNet struggles with anterior and posterior commissure alignment across contrasts, a step that the method prevents users from disabling.

Fig.~\ref{fig:runtime} compares end-to-end CPU runtimes for T1w-only and T1w-T2w inputs, using a single computational thread. On average, Infant-SynthSeg takes 56 minutes per T1w-T2w pair, most spent on its postprocessing routines. BIBSNet requires roughly half that time, but still needs over 10 minutes even for a single modality. In contrast, \method{} runs in about 2 minutes, providing more than an 11-fold speed-up over BIBSNet and a 24-fold speed-up over Infant-SynthSeg. On the GPU, \method{} takes 2 seconds per image, with 20 seconds one-time setup -- equivalent to 1,800 segmentations per hour.

\method{} has several limitations. It does not delineate the vermis, ventricular subtypes, or cerebrospinal fluid. It also currently assumes prior registration of the input images. Simulating alignment errors within the data engine or predicting a separate label map for each input may mitigate this constraint. Finally, our implementation processes all inputs in parallel, which limits segmentation of subjects with many modalities to systems with sufficient memory.

\section{Conclusion}

We present \method{}, a novel brain segmentation framework for infants and young children. We integrate a powerful group convolutional mechanism with a training strategy that incorporates real and synthetic images of tremendous variability. This enables the method to leverage features from a flexible number of inputs of any MRI contrasts and resolutions without retraining. Our experiments demonstrate state-of-the-art performance, from extremely preterm neonates across postnatal development up to five years of age, using a single model.

By providing a fast and versatile segmentation framework, \method{} supports flexible experimentation with regards to the inputs provided. \method{} is available at \website{}.

\section*{Acknowledgment}

The authors thank Istv\'an N.\ Husz\'ar for sharing insights into phenotype-informed sampling, S.\ Mazdak Abulnaga for discussing neural network implementations, Jingru Fu for software testing, and Nathan X.\ Ngo and Ashvath Balgovind for data curation.
M.\ H.\ maintains a consulting relationship with Neuro42, A.\ V.\ D.\ with DeepHealth and Radence. Massachusetts General Hospital and Mass General Brigham manage these interests under their conflict-of-interest policies.

\bibliographystyle{plain}
\bibliography{main}

\end{document}